\pdfoutput=1
\documentclass[11pt]{article}
\usepackage{acl2023}
\usepackage{times}
\usepackage{latexsym}
\usepackage[T1]{fontenc}
\usepackage[utf8]{inputenc}
\usepackage{microtype}
\usepackage{inconsolata}
\usepackage{booktabs}
\usepackage{graphicx}
\graphicspath{{figures/notebook_outputs/}}
\newif\ifshowsupplement
\showsupplementfalse 
\title{Vectorizing Classical Tamil: Representation Learning for Verse--Commentary Pairs}

\author{
  Amrit Gopinath$^{1}$, Sangeetha Sivanesan$^{2}$ \\
  $^{1}$Sri Sivasubramaniya Nadar College of Engineering, Chennai, India \\
  $^{2}$National Institute of Technology Tiruchirappalli, Tamil Nadu, India \\
  $^{1}$Email: \texttt{amrit2410182@ssn.edu.in}
}

\begin{document}
\maketitle

\begin{abstract}
We construct a corpus of 1,262 verse--commentary (\textit{urai}) pairs from
five Classical Tamil source sections, ranging from technical grammatical prose
to modern paraphrase, and ask what information representation learning can
recover. We train recurrent and Transformer encoders, a Siamese-style
pair-matching network, an mBART-style encoder--decoder, and a decoder-only
language model. Each analysis is interpreted against an appropriate control on
the same data. TF-IDF provides a strong no-training lexical retrieval baseline,
alongside representation analyses and generation controls for the learned
models. A fixed string containing the 25 most frequent commentary words scores
higher on generation overlap than the decoder-only model. Canonical correlation
reaches 1.000 on Gaussian noise at these sample
sizes, token-F1 spans only about 0.02--0.20 on this corpus, and the
encoder--decoder continues to lower training loss for sixteen epochs after
validation loss has begun to rise. One narrow result remains: the decoder-only
model prefers authentic word order in 107 of 112 minimal-pair comparisons
(95.5\%), but does not reproduce held-out commentary content. We release the
extraction and evaluation protocol; redistribution of the source commentaries
remains subject to permission.
\end{abstract}

\section{Introduction}
\label{sec:introduction}

Throughout this paper, \textit{urai} refers to the commentary associated with a verse.

Classical Tamil has a long textual and commentary tradition, but it remains largely absent from contemporary NLP. Tamil-capable models are trained mainly on Modern Tamil and web text, including the AI4Bharat/IndicNLP corpus \citep{kunchukuttan2020indicnlp} and multilingual models such as mBERT \citep{devlin2019bert}. Modelling Classical Tamil is therefore a small-data problem in its own right, not a smaller version of modern-Tamil modelling.

We collected 1,262 verse--urai pairs from the Tamil Virtual University \citep{tamilvu} and Project Madurai \citep{projectmadurai}. The corpus combines three explanatory registers: a modern paraphrase, gloss-and-gist material, and sectioned scholarly commentary. This variation matters because the same verse-to-urai task has different target formats across sources.

Our question is straightforward: when a model is trained on verse--urai pairs, does it learn anything beyond lexical overlap and recurring commentary forms? We examine recurrent and attention-based encoders, encoder--decoder generation, and decoder-only continuation. Cosine similarity, CCA, KCCA, and t-SNE provide views of the learned spaces; separate verse and urai encoders are compared with shared and Siamese-style settings.

The encoder--decoder lets us inspect how a verse is mapped to an urai and whether high attention weights reflect more than repeated vocabulary. We also test generalisation and use a pair-choice probe in which words or clauses are modified and reordered. That probe measures probability preference for the authentic order; it does not demonstrate grammatical understanding.

This is not a study of a deployable generator. It is a pilot study of a controls-first protocol for identifying what small-data models learn from verse--urai pairs, and what they do not.

\section{Related Work}
\label{sec:related-work}

Tholkappiyam is an early Tamil grammatical work with an extensive commentary tradition \citep{zvelebil1975tamil}. Its text is available digitally through Project Madurai \citep{projectmadurai} and the Tamil Virtual University \citep{tamilvu}, but these editions do not by themselves provide a consistently aligned verse--urai corpus. Constructing such pairs requires both textual extraction and explicit decisions about where a verse, its commentary, and a usable pair begin and end.

Classical-language NLP is further developed for Sanskrit, with a dependency treebank, specialised tools for segmentation, tagging, parsing, and annotation, and pretrained models evaluated on Sanskrit tasks \citep{kulkarni2020sanskrittreebank,sandhan2023sanskritshala,nehrdich2024byt5sanskrit}. Large Indian-language resources and multilingual models instead focus on modern text \citep{kunchukuttan2020indicnlp,kakwani2020indicnlpsuite,dabre2022indicbart}. Their existence does not establish transfer to a small historical corpus. Work on languages unseen during multilingual pretraining likewise shows that adaptation and evaluation matter beyond the presence of a multilingual model \citep{muller2021unseen}. This gap motivates in-domain data and careful controls.

We use paired-representation methods such as Siamese networks and CCA \citep{bromley1993siamese,hotelling1936cca,hardoon2004kcca}, and a word-order probe grounded in Old Tamil regularities \citep{herring2000poeticality}. Recoverability is not evidence that a model uses a linguistic property \citep{belinkov2022probing}; our evaluation therefore distinguishes alignment, generation, and probability preference.

\section{Dataset Construction}
\label{sec:dataset}

\subsection{Sources}

We construct a verse--commentary (\textit{urai}) corpus from five Classical Tamil sources in two digital libraries: Project Madurai \citep{projectmadurai} and the Tamil Virtual University \citep{tamilvu}. We parsed Naaladiyar from Project Madurai's Unicode text and used its trailing numerical markers (1--400) to recover verse--urai pairs. Thirukadukam and the three Tholkappiyam adhikarams used here -- Ezhuthatikaram, Sollathikaram, and Porulathikaram -- were obtained from the Tamil Virtual University. The Tholkappiyam material was collected manually rather than through an automated crawler. We normalize all sources into JSONL records with \texttt{verse}, \texttt{urai}, and \texttt{dataset} fields.

\subsection{Filtering}

We drop rows with an empty verse or urai after parsing. This leaves 1,262 usable pairs from 1,277 parsed records. No additional length-based filtering is applied in the v3 experiments.

\subsection{Statistics}

Table~\ref{tab:dataset-stats} reports the number of valid verse--urai pairs used for the models after filtering.

\begin{table}[h]
\centering
\small
\begin{tabular}{llr}
\toprule
\textbf{Source} & \textbf{Urai register} & \textbf{Pairs} \\
\midrule
Naaladiyar                    & Modern paraphrase & 393 \\
Tholk.\ Elutt.                & Scholarly commentary & 379 \\
Tholk.\ Soll.                 & Scholarly commentary & 287 \\
Tholk.\ Porul.                & Scholarly commentary & 103 \\
Thirukadukam                  & Gloss/gist/notes & 100 \\
\midrule
\textbf{Total} & & \textbf{1{,}262} \\
\bottomrule
\end{tabular}
\caption{Verse--urai pairs used for training, counted at load time after
filtering (Section~\ref{sec:dataset}). Thirukadukam's commentator and date
remain unestablished; the Naaladiyar material is a modern paraphrase, not a
traditional commentary.}
\label{tab:dataset-stats}
\end{table}

The resulting corpus contains 1,262 verse--urai pairs.

Urai are substantially longer than their verses on average, as expected for prose explanations of terse, often aphoristic text. The gap is largest in Eluttatikaram, where short phonological sutras receive extended commentary.

\section{Experimental Setup and Methodology}
\label{sec:methodology}

\subsection{Controls before models}

With 1,262 pairs, a score is interpretable only against a matched control. We use TF-IDF retrieval as a no-training lexical baseline, Gaussian matrices with the same $n$ and dimensionality for CCA, and a token-F1 null distribution from unigram sampling, random training urai, verse copying, a constant 25-word string, and a training-set oracle. TF-IDF provides a no-training lexical retrieval reference, while the learned-representation analyses reported below use different diagnostics and are not directly comparable as retrieval evaluations. The Gaussian control tests whether CCA is high because the problem is underconstrained. The token-F1 nulls show what overlap is attainable without conditioning on the verse.

\subsection{Models}

All models are trained from scratch on the corpus, with seed 3407.

We train separate LSTM, BiLSTM, and small Transformer encoders with a masked-word objective, extracting last-state and mean-pooled sentence vectors. We compare them with a Siamese Transformer trained to assign 1 to matched pairs and 0 to mismatches. For generation, we train a 256-dimensional, 6+6-layer mBART-style encoder--decoder with denoising followed by verse-to-urai fine-tuning, and a 256-dimensional, 4-layer decoder-only model trained on verse language modelling and verse-to-urai continuation.

\subsection{Tokenisation}

We tokenise on whitespace and keep whole words. Classical Tamil is agglutinative, so this is a limitation, and we return to it in Section~\ref{sec:limitations}.

All reported experiments use the v3 token set: \texttt{<pad> <unk> <mask> <bos> <eos> <verse> <urai>}.

\subsection{Splits and evaluation}

We use a deterministic 90/10 split: 1,136 training pairs and 126 held-out pairs. Generation and token-F1 use held-out verses only. The vocabulary is built on the full corpus, so we do not evaluate out-of-vocabulary words. The TF-IDF baseline is evaluated on the full corpus and reported by source because word-gloss commentary and paraphrase produce very different amounts of lexical overlap. No learned retrieval result is reported.

\subsection{Grammar probe}

Generation alone cannot determine whether the decoder prefers the word order of an original verse: a generated urai can differ from its reference in words, length, and meaning. We therefore use a pair-choice test. Each item contains a real verse and an equal-length version with one controlled perturbation, so the comparison is not driven by sequence length.

The 113 constructed pairs come from ten Naaladiyar and Porulathikaram verses. Thirteen target three regularities from \citet{herring2000poeticality}; 99 randomly reorder one line, and one changes an inflection. The inflection item is excluded from the pooled word-order result because its perturbation contains an out-of-vocabulary token. We sum token log-probabilities and count a pair as correct when the original scores higher (chance 0.5). The set is small, dominated by random reorderings, and based on partly hand-checked language-model parses. It is therefore a diagnostic of local preference, not a measure of broad grammatical competence.

\section{Results and Analysis}
\label{sec:results}

\subsection{What the controls achieve}

TF-IDF retrieves the correct urai at rank 1 in 35.3\% of cases (Table~\ref{tab:tfidf}), whereas chance is below 1\%. Thirukadukam reaches 92.0\% because its urai repeats verse words as gloss headwords; lexical retrieval is nearly solved there by construction. Eluttatikaram and Sollathikaram reach 28.0\% and 32.1\%, while Porulathikaram reaches 54.4\%. TF-IDF is therefore the relevant baseline, not chance.

\begin{table}[h]
\centering
\small
\begin{tabular}{lrrr}
\toprule
\textbf{Pool} & \textbf{n} & \textbf{R@1} & \textbf{MRR} \\
\midrule
Naaladiyar                   & 393   & 0.389 & 0.472 \\
Tholk.\ Eluttatikaram        & 379   & 0.280 & 0.382 \\
Tholk.\ Sollatikaram         & 287   & 0.321 & 0.402 \\
Tholk.\ Porulatikaram        & 103   & 0.544 & 0.614 \\
Thirukadukam                 & 100   & 0.920 & 0.945 \\
\midrule
Combined                     & 1{,}262 & 0.353 & 0.429 \\
\bottomrule
\end{tabular}
\caption{TF-IDF lexical retrieval, no training. Chance R@1 is below 0.01 in
every pool. Thirukadukam is close to solved lexically because its commentary
glosses the verse word by word.}
\label{tab:tfidf}
\end{table}

\subsection{Canonical correlation is not interpretable at this sample size}

The vectors are 256-dimensional, while pools contain 100--393 pairs. CCA on matched Gaussian noise reaches 1.000 at both $n=100$ and $n=393$. Reducing the data to 32 dimensions with PCA lowers these values only to 0.905 and 0.542. Raw CCA is not evidence of alignment in this setting; it must be interpreted against a random-data baseline with the same sample size and dimensionality.

\subsection{Token-F1 requires a null baseline}

On held-out rows, the decoder scores 0.060 token-F1, below TF-IDF retrieval (0.084) and a constant 25-word string (0.108; Table~\ref{tab:f1null}). The oracle is 0.195, leaving a narrow usable range of roughly 0.02--0.20. An identical rerun produced 0.057, so we do not interpret differences below about 0.02.

\begin{table}[h]
\centering
\small
\begin{tabular}{lr}
\toprule
\textbf{System} & \textbf{token-F1} \\
\midrule
Unigram-sampled words            & 0.022 \\
Random training urai             & 0.030 \\
Verse copied as its own urai     & 0.044 \\
Random urai from the same text   & 0.058 \\
\textbf{Decoder-only model}      & \textbf{0.060} \\
TF-IDF retrieved urai            & 0.084 \\
\textbf{Constant top-25 words}   & \textbf{0.108} \\
\midrule
Oracle (best training urai)      & 0.195 \\
\bottomrule
\end{tabular}
\caption{Null distribution for token-F1 on the same 30 held-out rows. The
trained model scores below a fixed string that ignores the input, and below
lexical retrieval. The oracle bounds what the metric can express on this
corpus.}
\label{tab:f1null}
\end{table}

\subsection{Separate encoders do not recover verse--urai alignment}

The verse and urai encoders are trained separately, with no objective requiring a verse to be close to its corresponding commentary. We therefore test whether matched pairs are close in the learned vector spaces. They are not. For LSTM last-state vectors, mean cosine similarity for matched pairs is 0.042 for Naaladiyar, 0.021 for Thirukadukam, and 0.045 for Tholkappiyam; it is 0.002 for the combined corpus. Mean-pooled vectors show the same pattern, with values from $-0.042$ to 0.018.

A small pair classifier also does not reliably distinguish a real pair from a random mismatch within a source. Its accuracy ranges from 0.500 to 0.600 across the individual source pools, where chance is 0.500. When all sources are pooled, accuracy rises to 0.842--0.845. This pooled result is not evidence of verse--urai alignment: negatives are sampled from the full corpus, so the classifier can use source-specific vocabulary, length, and commentary register instead of the relationship between a particular verse and its urai.

\begin{table}[h]
\centering
\scriptsize
\begin{tabular}{lcc}
\toprule
\textbf{Pool} & \textbf{Cosine} & \textbf{Classifier} \\
\midrule
Naaladiyar & 0.042 / $-0.017$ & 0.509 / 0.500 \\
Thirukadukam & 0.021 / 0.008 & 0.500 / 0.505 \\
Tholkappiyam & 0.045 / 0.018 & 0.554 / 0.600 \\
All sources & 0.002 / $-0.042$ & 0.842 / 0.845 \\
\bottomrule
\end{tabular}
\caption{Separate-encoder diagnostics. Each entry is last-state / mean-pooled.
Cosine values are mean similarities for matched verse--urai pairs. Classifier
results use randomly mismatched pairs; the pooled classifier can exploit source
identity.}
\label{tab:separate-encoders}
\end{table}

\subsection{t-SNE and Siamese matching show what the objective encourages}

We use t-SNE only as a visual diagnostic. In the combined representation plots, points group more clearly by source text and commentary register than by their individual verse--urai pairing. This is consistent with the classifier result: the learned representations retain broad corpus distinctions, but the plots do not show that a verse is placed near its own urai. Since t-SNE reduces high-dimensional vectors to two dimensions and can exaggerate apparent clusters, it supports the quantitative results rather than serving as evidence on its own.

The Siamese attention model gives a different result because it is trained directly to separate matched from mismatched pairs. On the same pairs used for training, the mean cosine gap between a real pair and a random mismatch is 0.517 for Naaladiyar, 0.462 for Thirukadukam, and 0.500 for Tholkappiyam; the combined model has a smaller gap of 0.345. This shows that an explicit matching objective can fit the observed pair labels. It does not establish generalisation, because the model is evaluated on training pairs and the negative pairs are randomly sampled. A held-out, within-source retrieval or pair-classification test is needed before treating this as evidence of general verse--urai alignment.

\begin{figure*}[ht]
\centering
\includegraphics[width=.47\textwidth]{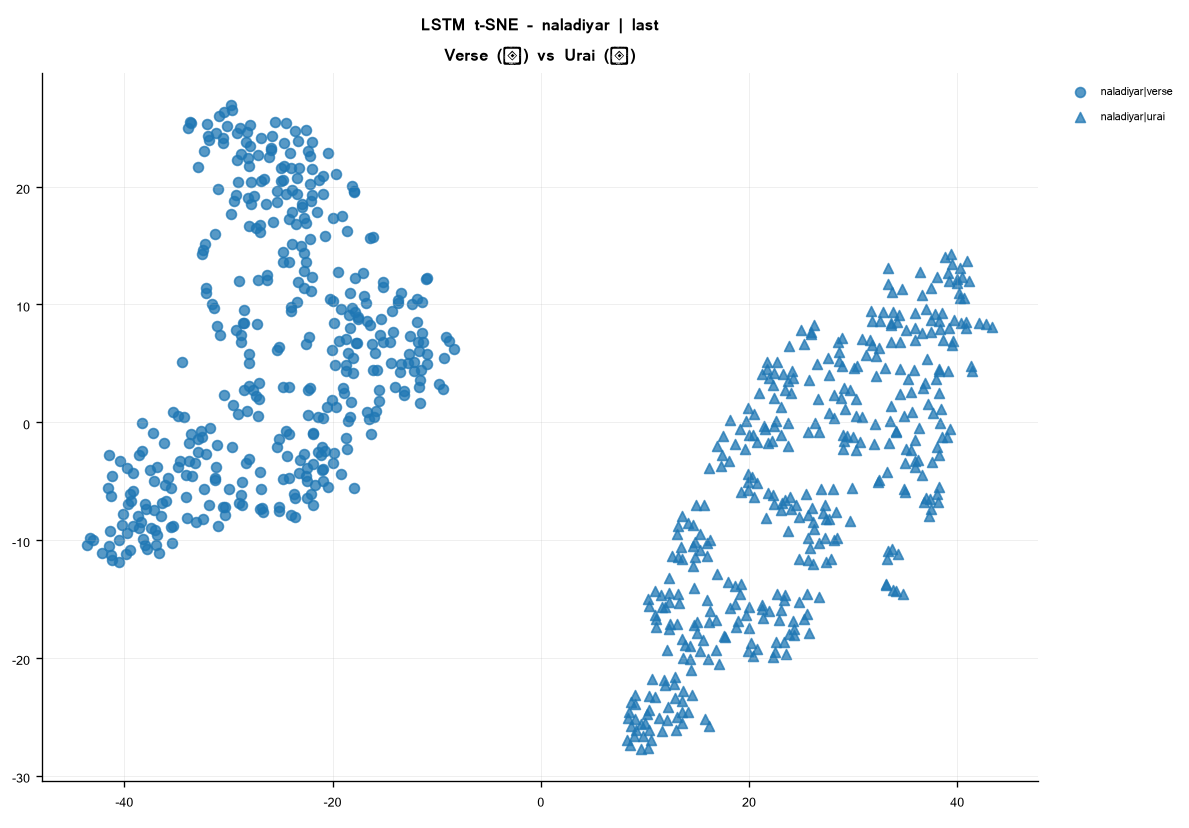}\hfill
\includegraphics[width=.47\textwidth]{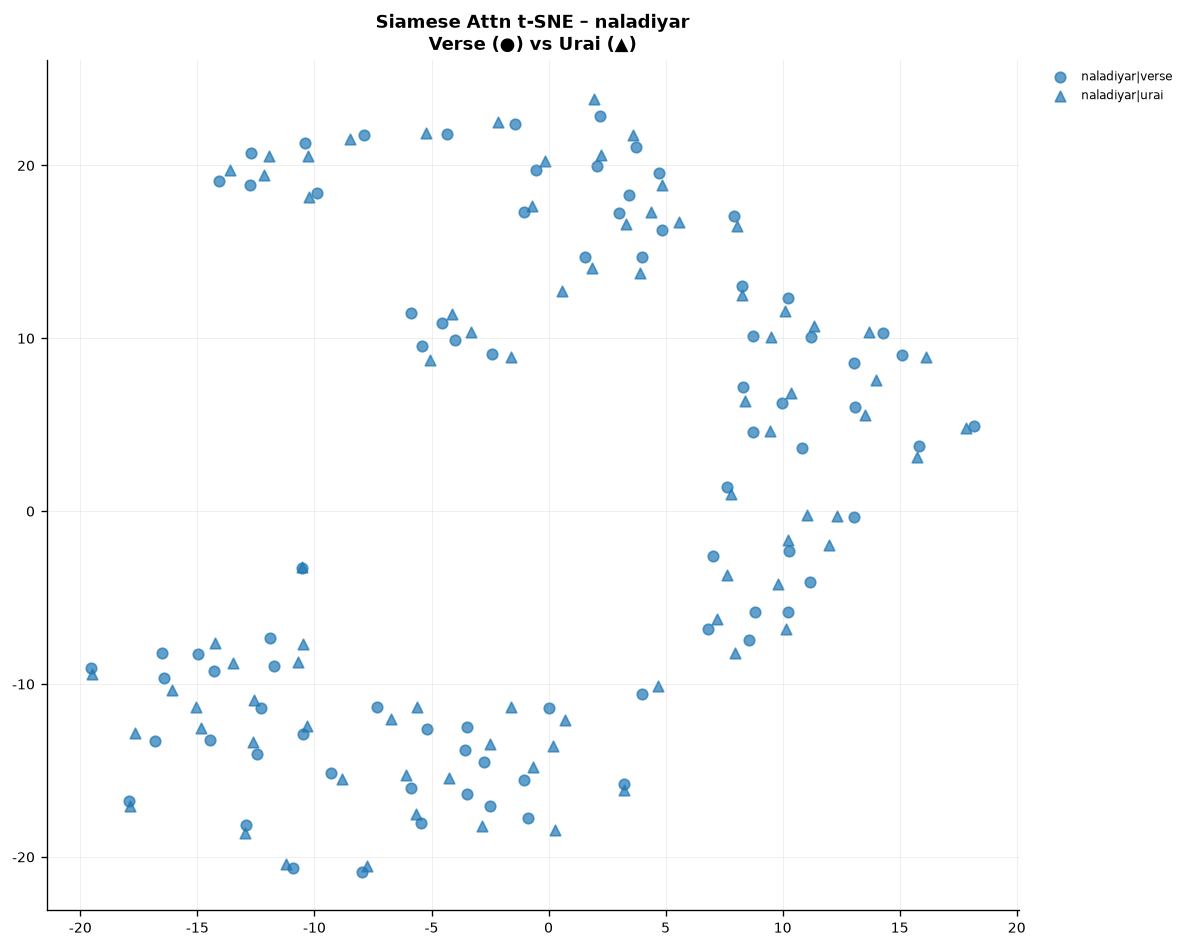}
\par\smallskip
{\small \(\bullet\) Verse representation \qquad \(\triangle\) Urai representation}
\caption{t-SNE diagnostics for Naaladiyar representations. In both panels, the horizontal axis is t-SNE dimension~1 and the vertical axis is t-SNE dimension~2; the coordinates are arbitrary projection coordinates, not interpretable linguistic features. Circles denote verse representations and triangles denote commentary representations. \textbf{Left:} separately trained LSTM verse and commentary encoders occupy distinct regions. \textbf{Right:} the Siamese model places many training verse--commentary pairs near one another because its objective explicitly rewards pair matching. These two-dimensional projections are not quantitative evidence of alignment or generalisation.}
\label{fig:tsne-diagnostic}
\end{figure*}

Figure~\ref{fig:tsne-diagnostic} provides a visual contrast between these two settings for Naaladiyar. The projection is included to show what the objectives encourage, not as an additional alignment test.

\begin{figure*}[ht]
\centering
\includegraphics[width=.95\textwidth]{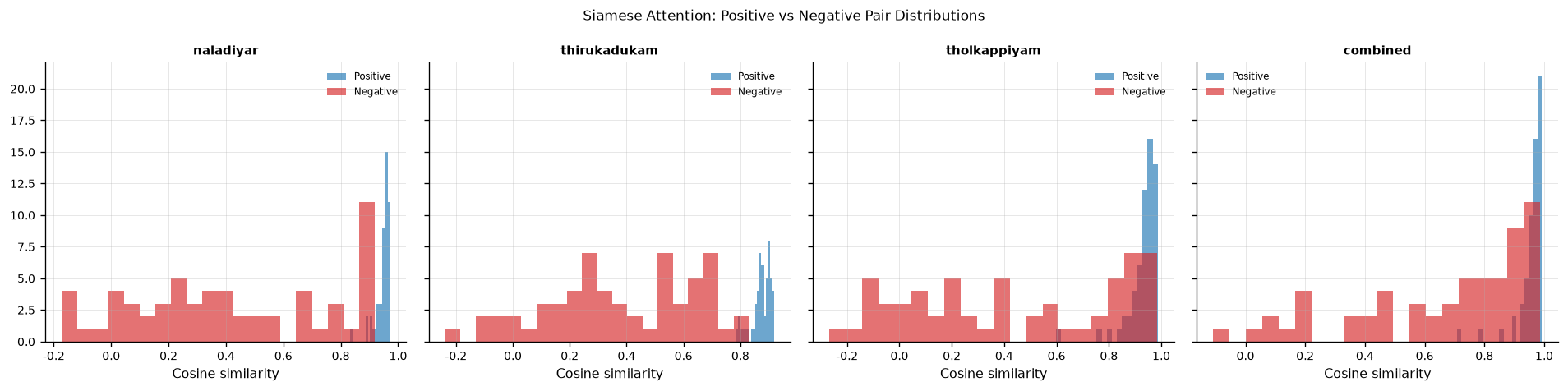}
\caption{Siamese attention scores for matched (positive) and randomly
mismatched (negative) training pairs. The matching objective separates these
labels within each pool; the plot is not a held-out generalisation result.}
\label{fig:siamese-distributions}
\end{figure*}

\subsection{The encoder--decoder overfits after epoch 4}

\begin{figure*}[ht]
\centering
\includegraphics[width=.95\textwidth]{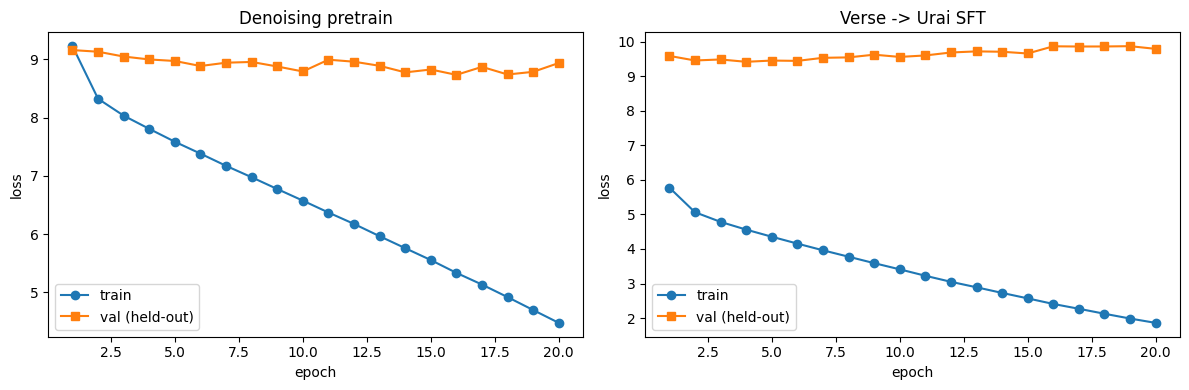}
\caption{Training and held-out loss for the two encoder--decoder stages. The
supervised validation loss is lowest at epoch 4, while training loss continues
to fall through epoch 20.}
\label{fig:loss-curves}
\end{figure*}

During supervised fine-tuning, training loss falls from 5.78 to 1.86. Held-out loss reaches its minimum at epoch 4 (9.42) and rises to 9.79 by epoch 20 (Table~\ref{tab:curves}; Figure~\ref{fig:loss-curves}). We therefore restore the epoch-4 checkpoint. The epoch-20 training loss reflects memorisation rather than improved generalisation.

\begin{table}[h]
\centering
\small
\begin{tabular}{lrrr}
\toprule
\textbf{Stage} & \textbf{train} & \textbf{val} & \textbf{val min} \\
\midrule
Denoising & 9.23 $\rightarrow$ 4.47 & 9.16 $\rightarrow$ 8.94 & 8.73 (ep.\ 16) \\
Fine-tune & 5.78 $\rightarrow$ 1.86 & 9.59 $\rightarrow$ 9.79 & 9.42 (ep.\ 4) \\
\bottomrule
\end{tabular}
\caption{Encoder--decoder loss over 20 epochs per stage. In the fine-tuning
stage validation loss reaches its minimum at epoch 4 and is worse at epoch 20
than at epoch 1, while training loss continues to fall.}
\label{tab:curves}
\end{table}

\subsection{Generated commentary reuses training material}

We inspected five held-out generations to assess whether the decoder produced input-specific commentary or reused training material. Four outputs begin with the same phrase despite being generated for different Naaladiyar verses. The inspection also found long exact phrases from the training corpus, often from rows unrelated to the input verse. These examples suggest that the decoder combines familiar commentary fragments rather than producing an explanation specific to the verse.

The same pattern appears on out-of-domain Thirukkural couplets. The model produces outputs with a mean length of 102.4 tokens, but the mean token overlap between the input couplet and the generated output is 0.000. It therefore continues to produce commentary-like text without using vocabulary from the verse it is asked to explain.

\subsection{Cross-attention recovers vocabulary overlap}

Across 579,818 word pairs, the top twenty attention weights all come from Tholkappiyam; half link a word to the same word reused in the urai. The strongest attention patterns therefore reflect verbatim vocabulary reuse, not semantic alignment beyond TF-IDF.

\begin{table}[h]
\centering
\small
\begin{tabular}{lr}
\toprule
\textbf{Top-attention property} & \textbf{Count} \\
\midrule
Eluttatikaram pairs & 16 / 20 \\
Sollathikaram pairs & 3 / 20 \\
Porulathikaram pairs & 1 / 20 \\
Exact verse--urai word matches & 10 / 20 \\
\bottomrule
\end{tabular}
\caption{Composition of the twenty highest cross-attention word pairs. All
twenty come from Tholkappiyam.}
\label{tab:attention-top20}
\end{table}

\subsection{Form is learned where content is not}

The decoder-only model prefers the original order in 107/112 pairs (95.5\%), but its held-out token-F1 is 0.060, below the 0.108 constant baseline. It captures a preference for form without recovering commentary content. Preference is not evidence of use \citep{belinkov2022probing}, and an unseen-bigram control is still needed to separate grammatical sensitivity from local distributional cues.

\begin{figure}[ht]
\centering
\includegraphics[width=\linewidth]{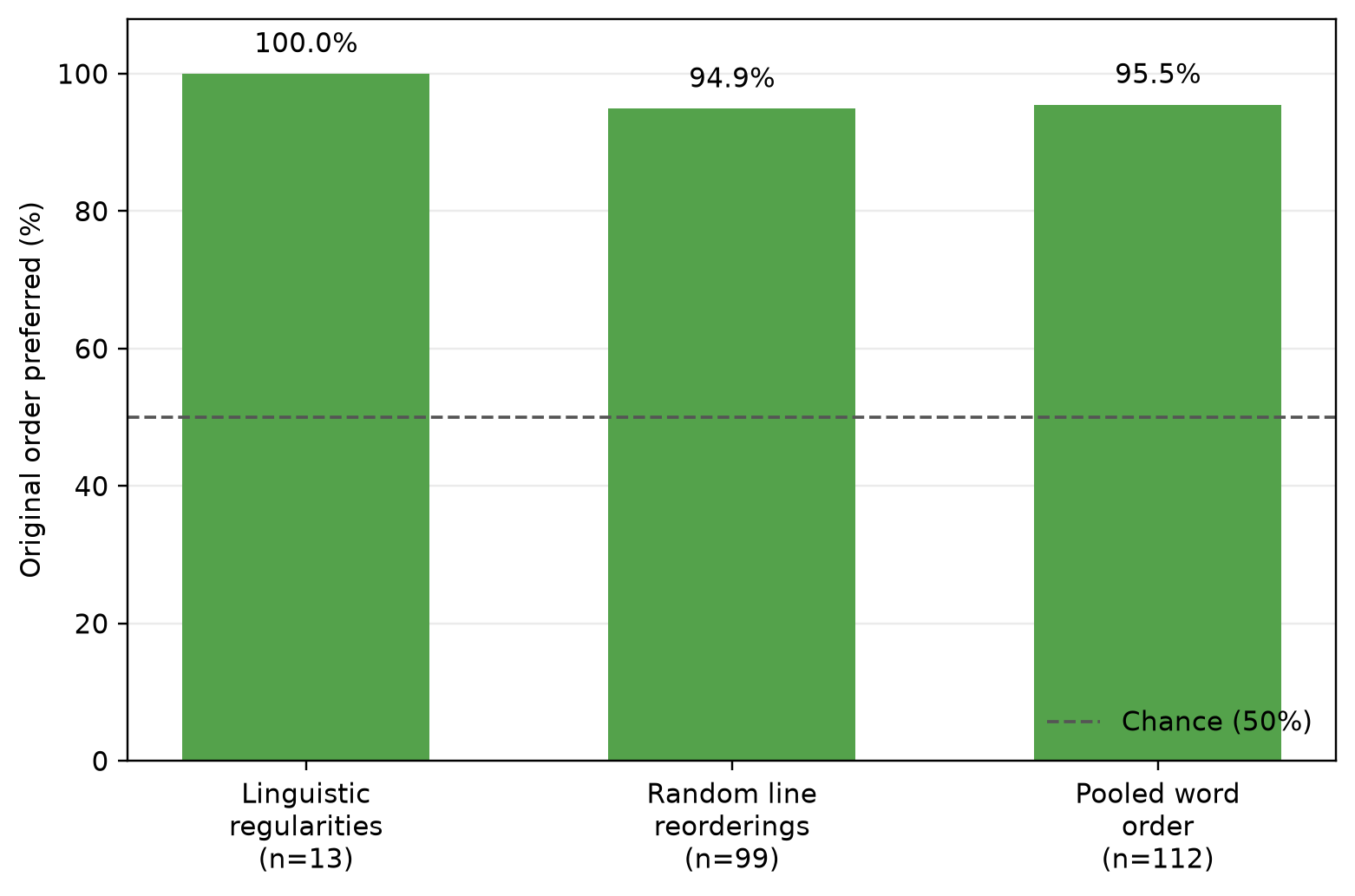}
\caption{Final v3 minimal-pair results. The inflection item is excluded because
its perturbation contains an out-of-vocabulary token. The 95.5\% pooled result
is driven primarily by 99 random line reorderings.}
\label{fig:grammar-probe-final}
\end{figure}

Figure~\ref{fig:grammar-probe-final} separates the small set of targeted regularities from the random line reorderings. The high pooled score should be read alongside the imbalance between these subsets.

\subsection{The input does not specify the target register}

The explanatory styles also differ sharply in length: urai are about 1.7$\times$ longer than verses in Naaladiyar, 3.2$\times$ longer in Thirukadukam, and 5.2--8.7$\times$ longer across the Tholkappiyam sections. A verse does not reliably indicate which register it requires. A model may generate scholarly commentary-style language for a Naaladiyar verse even though its target is a modern paraphrase. Future work should either provide the target register as an input or train separate models for each register, then evaluate whether the output matches both the verse content and the intended explanatory style.

\begin{figure*}[t]
\centering
\includegraphics[width=.88\textwidth]{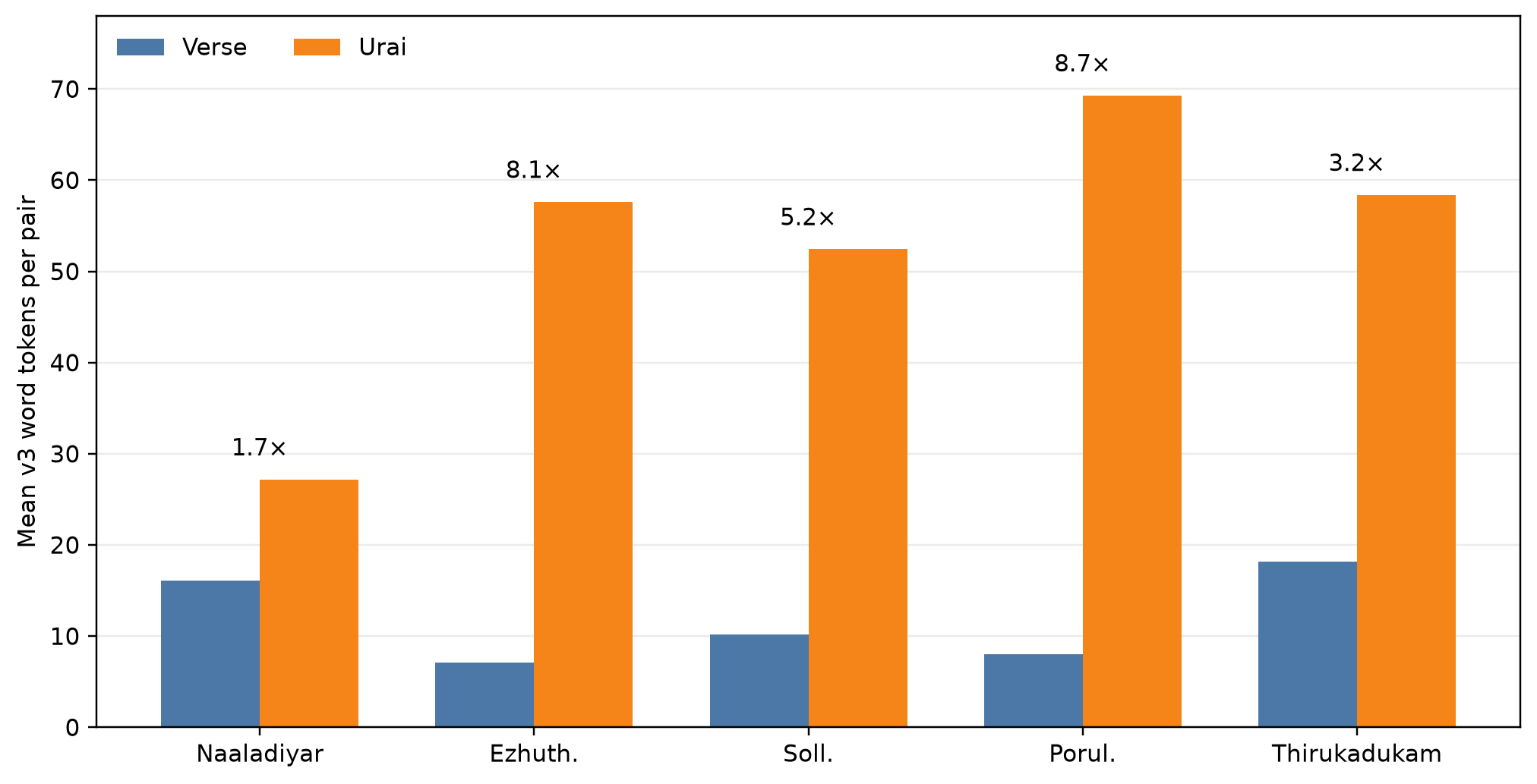}
\caption{Mean verse and urai lengths under the v3 word tokeniser. The ratio
above each source is mean urai length divided by mean verse length. The target
format differs substantially by source before a model sees the input verse.}
\label{fig:register-length}
\end{figure*}

\section{Reproducibility}
\label{sec:reproducibility}
The anonymous repository contains output-stripped source notebooks for the reported controls, LSTM, BiLSTM/Siamese, encoder--decoder, and decoder-only grammar-probe analyses. All five notebooks have unexecuted code cells and no saved cell outputs. The release also includes the static PNG figure assets used in this paper, but these are provided as figures rather than as embedded executed-notebook outputs.

The original source texts, derived JSONL records, trained weights, execution logs, and executed notebooks are not  distributed. The reported encoder--decoder values were obtained from the executed Kaggle run; the corresponding public notebook is source-only and can be inspected or rerun after independently obtaining the source material and preparing records with the \texttt{verse}, \texttt{urai}, and \texttt{dataset} fields.

The repository includes parsers for a locally obtained Project Madurai Naaladiyar text and for Tamil Virtual University Thirukadukam pages, together with a source map and the applicable attribution and academic-use notes. The Tholkappiyam material was manually collected and aligned, so no automated parser is supplied for that portion. Users must obtain all source material directly from the named libraries, follow their current terms of use, and seek permission before redistributing source or derived text. Anonymous code and reproduction materials are available at \url{https://anonymous.4open.science/r/Vectorisation-Classical-Tamil-NLP-7E69/}

\section{Limitations}
\label{sec:limitations}

The corpus contains 1,262 usable verse--urai pairs from five source sections representing three major works: Naaladiyar, Tholkappiyam, and Thirukadukam. Under the reported whitespace tokenisation, it contains 75,099 word tokens: 14,576 verse tokens and 60,523 urai tokens. It is small and uneven: Porulathikaram contributes 103 pairs, whereas Naaladiyar contributes 393.

Each model is evaluated with one configuration and seed. The train--held-out split contains verses from the same source collections on both sides, so it does not test transfer to a new text, commentator, period, or editorial tradition. Token-F1 is evaluated on 30 held-out rows rather than the complete held-out split. The vocabulary is built from the full corpus before splitting to avoid unknown-token failures; consequently, the experiments do not test out-of-vocabulary generalisation.

The representation and Siamese analyses are measured on training pairs. Their negative examples are random mismatches, and pooled mismatches may come from another source text. A high score can therefore reflect source identity, length, vocabulary, or commentary register rather than alignment between a verse and its own urai. t-SNE is likewise a qualitative projection, not a statistical test of clustering or alignment. Repeated seeds and held-out, within-source retrieval are needed before making stronger claims about representation learning.

Whole-word tokenisation is a poor fit for agglutinative Tamil: a meaningful stem or suffix change can receive zero whole-word overlap while preserving much of the intended content. The preprocessing also does not retain verse line structure. Work on metre or prosody will need a representation that handles line structure consistently across original and perturbed text.

The grammar probe is narrow. It contains 112 usable word-order pairs and is dominated by random line reorderings. Some parses were machine-generated and only partly checked by hand. Although the model assigns higher probability to the original string under these alterations, this does not establish broad grammatical competence or show that the preference is used when generating urai.

Finally, urai are not uniform ground-truth labels. The corpus combines modern paraphrase, lexically overlapping gloss material, and scholarly commentary. These are historically and generically different ways of explaining a verse. A plausible output in one style may be inappropriate in another, even if it has lexical overlap with the gold text. The paper does not include independent expert adjudication of every extracted pair or generated explanation.

\section{Conclusion}
\label{sec:conclusion}

No held-out learned verse--urai retrieval result is reported for this 1,262-pair corpus. TF-IDF establishes a strong lexical retrieval baseline, while the representation and Siamese analyses are non-comparable diagnostics; a constant string outperforms the decoder-only model on generation overlap. The study instead identifies conditions for interpreting small-data results: CCA requires matched noise controls, token-F1 requires a null distribution, compared inputs must be tokenised identically, and epoch selection requires held-out curves.

The decoder prefers authentic order in 95.5\% of minimal pairs while failing to reproduce commentary content. That separation between form and content is a diagnostic, not evidence of grammatical competence.

\bibliography{references}
\bibliographystyle{acl_natbib}

\end{document}